\documentclass[conference]{IEEEtran}
\IEEEoverridecommandlockouts

\usepackage{cite}
\usepackage{amsmath,amssymb,amsfonts}
\usepackage{algorithmic}
\usepackage{graphicx}
\usepackage{textcomp}
\usepackage{xcolor}
\usepackage{tikz}
\usepackage{forest}

\def\BibTeX{{\rm B\kern-.05em{\sc i\kern-.025em b}\kern-.08em
    T\kern-.1667em\lower.7ex\hbox{E}\kern-.125emX}}

\makeatletter 
\newcommand{\linebreakand}{%
  \end{@IEEEauthorhalign}
  \hfill\mbox{}\par
  \mbox{}\hfill\begin{@IEEEauthorhalign}
}
\makeatother 
    
\begin{document}

\title{Guided Code Generation with LLMs: A Multi-Agent Framework for Complex Code Tasks
}

\author{\IEEEauthorblockN{
Amr Almorsi\IEEEauthorrefmark{1}, 
Mohanned Ahmed\IEEEauthorrefmark{1}, 
Walid Gomaa\IEEEauthorrefmark{1}\IEEEauthorrefmark{2}}
\IEEEauthorblockA{
\IEEEauthorrefmark{1}\textit{ECCE School, Faculty of Engineering, Egypt-Japan University of Science and Technology}\\
\IEEEauthorrefmark{2}\textit{Faculty of Engineering, Alexandria University}\\
Alexandria, Egypt \\
Email: \{amr.saleh, mohanned.hafez, walid.gomaa\}@ejust.edu.eg}
}

\maketitle

\begin{abstract}
Large Language Models (LLMs) have shown remarkable capabilities in code generation tasks, yet they face significant limitations in handling complex, long-context programming challenges and demonstrating complex compositional reasoning abilities. This paper introduces a novel agentic framework for ``guided code generation'' that tries to address these limitations through a deliberately structured, fine-grained approach to code generation tasks. Our framework leverages LLMs' strengths as fuzzy searchers and approximate information retrievers while mitigating their weaknesses in long sequential reasoning and long-context understanding. Empirical evaluation using OpenAI's HumanEval benchmark with Meta's Llama 3.1 8B model (int4 precision) demonstrates a 23.79\% improvement in solution accuracy compared to direct one-shot generation. Our results indicate that structured, guided approaches to code generation can significantly enhance the practical utility of LLMs in software development while overcoming their inherent limitations in compositional reasoning and context handling.

\end{abstract}


\section{Introduction}

\par 
Large Language Models (LLMs) have revolutionized the field of code generation, offering unprecedented capabilities in automating software development tasks. These models have demonstrated impressive performance in simple code completion, generating code snippets, completing partial implementations, and even assisting in bug fixing and optimization. However, despite their remarkable achievements, current LLM-based code generation systems face significant limitations, particularly in planning, handling long, complex code tasks, and navigating large-scale coding projects.

\subsection{Limitations of Long-Context LLMs}

\par 
One of the primary challenges hindering the widespread adoption of LLMs for comprehensive code generation is the limited context window constraint. Most state-of-the-art LLMs are trained to operate within a fixed token limit, typically ranging from several hundred to a few thousand tokens. This restriction severely impedes the model's ability to ``reason'' over long sequences understand, and generate code for intricate software systems that often span thousands of lines across multiple files.

\par 
Nonetheless, recent breakthroughs in extending the context window massively up to 128k \cite{b1}\cite{b2}\cite{b3} have been made and even more recent approaches like Infini-attention\cite{infiniattention} and LongRoPE\cite{longrope} extend the context window to the order of hundreds of thousands and millions of tokens. However, when it comes to evaluating long-context LLMs, they are most commonly evaluated on needle-in-haystack\cite{needleinhaystack} tasks which test whether the LLM can successfully fetch pieces of information residing in a long sequence of tokens. However, this does not truly measure the LLMs' abilities to "reason" with these pieces of information in more realistic scenarios and practical tasks—like Coding, Reasoning, In-context learning, Multilingual comprehension, and others. For these reasons, LongICLBench\cite{longiclbench} has been compiled in order to evaluate long-context LLMs on in-context learning (ICL) on extreme-label classification tasks, which demonstrated that the performance of all state-of-the-art long-context LLMs degrades uniformly with respect to the length of the input.

\subsection{Limited Compositional Abilities of LLMs}

\par 
A significant limitation of Large Language Models (LLMs) lies in their compositional abilities and sequential reasoning capabilities. In a recent study, Xu et al. \cite{llmcompose} investigated LLMs' capacity to synthesize previously learned individual skills to address novel composite tasks. Through comprehensive empirical analyses across various LLM architectures, the researchers evaluated performance on a range of composite tasks encompassing both linguistic and logical challenges. Their findings revealed varying levels of compositional proficiency among LLMs, with models demonstrating competence in basic tasks but exhibiting notable difficulties with complex sequential reasoning problems. Through theoretical analysis in a simplified setting, the authors established that LLMs display compositional capabilities primarily when tasks can be effectively decomposed based on input parameters. While their investigation of model scaling effects indicated that larger models—with their more sophisticated internal architectures—showed enhanced performance on elementary composite tasks, a crucial limitation emerged: "for more complex composite tasks involving reasoning multiple steps, where each step represents one task, models typically underperform, \textit{and scaling up generally provides no improvements}" \cite{llmcompose}.

\par 
While contemporary applications of Large Language Models (LLMs) predominantly focus on code completion tasks (exemplified by GitHub Copilot), generation of discrete code segments, or environmental interaction capabilities as demonstrated in OpenHands \cite{openhands}, these implementations typically employ a single-step generation process. However, this approach presents a fundamental contradiction: computational processes are inherently sequential and compositional in nature, and considering the limitations in compositional reasoning identified by Xu et al. \cite{llmcompose}, the effectiveness of generating code—which fundamentally comprises sequential compositions of programming constructs—becomes theoretically questionable.

To address these limitations, we introduce a novel agentic framework for "guided code generation," which implements a more structured and granular approach to complex code generation tasks. This methodology capitalizes on LLMs' demonstrated capabilities as fuzzy searchers and approximate information retrievers while mitigating their compositional limitations. We evaluate our framework's efficacy using OpenAI's HumanEval benchmark, employing Meta's Llama 3.1 8B model with int4 precision\cite{nousresearchllama3.1} quantization due to computational constraints and limited resources. Additionally, we present a formal theoretical framework for analyzing code generation tasks within the context of LLMs and our proposed methodology.

\section{Proposed Method}

\par 
Our framework introduces a novel approach to structured code generation through a multi-agent system designed to break down complex coding tasks into manageable, composable units. The framework operates through three primary components: \textit{hierarchical decomposition}, \textit{bottom-up code generation}, and \textit{multi-agent validation}.

\subsection{Hierarchical Problem Decomposition}

\begin{figure}[ht]
\centering

\tikzset{every picture/.style={line width=0.75pt}} 

\begin{tikzpicture}[x=0.75pt,y=0.75pt,yscale=-1,xscale=1]

\draw (457.75,161.88) node  {\includegraphics[width=26.8pt,height=26.8pt]{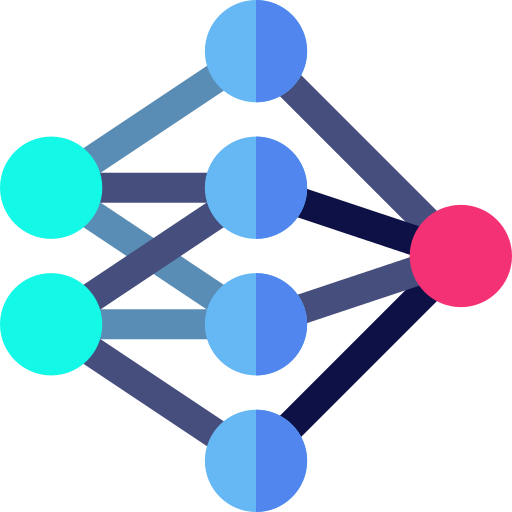}};
\draw (334.98,172.46) node  {\includegraphics[width=32.38pt,height=32.38pt]{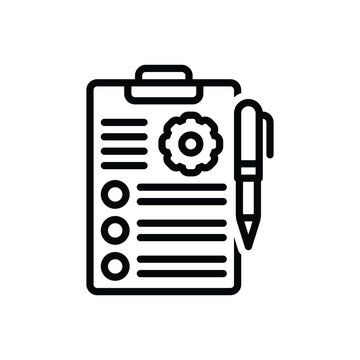}};
\draw   (571.92,153.09) .. controls (571.92,150.5) and (574.02,148.4) .. (576.61,148.4) .. controls (579.19,148.4) and (581.29,150.5) .. (581.29,153.09) .. controls (581.29,155.68) and (579.19,157.77) .. (576.61,157.77) .. controls (574.02,157.77) and (571.92,155.68) .. (571.92,153.09) -- cycle ;
\draw   (590.66,171.36) .. controls (590.66,168.77) and (592.76,166.68) .. (595.35,166.68) .. controls (597.94,166.68) and (600.03,168.77) .. (600.03,171.36) .. controls (600.03,173.95) and (597.94,176.05) .. (595.35,176.05) .. controls (592.76,176.05) and (590.66,173.95) .. (590.66,171.36) -- cycle ;
\draw   (571.92,171.36) .. controls (571.92,168.77) and (574.02,166.68) .. (576.61,166.68) .. controls (579.19,166.68) and (581.29,168.77) .. (581.29,171.36) .. controls (581.29,173.95) and (579.19,176.05) .. (576.61,176.05) .. controls (574.02,176.05) and (571.92,173.95) .. (571.92,171.36) -- cycle ;
\draw   (553.18,171.36) .. controls (553.18,168.77) and (555.28,166.68) .. (557.86,166.68) .. controls (560.45,166.68) and (562.55,168.77) .. (562.55,171.36) .. controls (562.55,173.95) and (560.45,176.05) .. (557.86,176.05) .. controls (555.28,176.05) and (553.18,173.95) .. (553.18,171.36) -- cycle ;
\draw   (604.72,190.1) .. controls (604.72,187.51) and (606.82,185.42) .. (609.4,185.42) .. controls (611.99,185.42) and (614.09,187.51) .. (614.09,190.1) .. controls (614.09,192.69) and (611.99,194.79) .. (609.4,194.79) .. controls (606.82,194.79) and (604.72,192.69) .. (604.72,190.1) -- cycle ;
\draw   (557.86,190.1) .. controls (557.86,187.51) and (559.96,185.42) .. (562.55,185.42) .. controls (565.14,185.42) and (567.24,187.51) .. (567.24,190.1) .. controls (567.24,192.69) and (565.14,194.79) .. (562.55,194.79) .. controls (559.96,194.79) and (557.86,192.69) .. (557.86,190.1) -- cycle ;
\draw   (585.98,190.1) .. controls (585.98,187.51) and (588.07,185.42) .. (590.66,185.42) .. controls (593.25,185.42) and (595.35,187.51) .. (595.35,190.1) .. controls (595.35,192.69) and (593.25,194.79) .. (590.66,194.79) .. controls (588.07,194.79) and (585.98,192.69) .. (585.98,190.1) -- cycle ;
\draw    (565.58,186.39) -- (572.39,176.78) ;
\draw [shift={(573.55,175.15)}, rotate = 125.31] [color={rgb, 255:red, 0; green, 0; blue, 0 }  ][line width=0.75]    (6.56,-1.97) .. controls (4.17,-0.84) and (1.99,-0.18) .. (0,0) .. controls (1.99,0.18) and (4.17,0.84) .. (6.56,1.97)   ;
\draw    (587.6,186.39) -- (580.37,176.75) ;
\draw [shift={(579.17,175.15)}, rotate = 53.13] [color={rgb, 255:red, 0; green, 0; blue, 0 }  ][line width=0.75]    (6.56,-1.97) .. controls (4.17,-0.84) and (1.99,-0.18) .. (0,0) .. controls (1.99,0.18) and (4.17,0.84) .. (6.56,1.97)   ;
\draw    (560.92,168.04) -- (571.4,158.02) ;
\draw [shift={(572.84,156.64)}, rotate = 136.28] [color={rgb, 255:red, 0; green, 0; blue, 0 }  ][line width=0.75]    (6.56,-1.97) .. controls (4.17,-0.84) and (1.99,-0.18) .. (0,0) .. controls (1.99,0.18) and (4.17,0.84) .. (6.56,1.97)   ;
\draw    (606.11,186.63) -- (599.07,176.78) ;
\draw [shift={(597.91,175.15)}, rotate = 54.46] [color={rgb, 255:red, 0; green, 0; blue, 0 }  ][line width=0.75]    (6.56,-1.97) .. controls (4.17,-0.84) and (1.99,-0.18) .. (0,0) .. controls (1.99,0.18) and (4.17,0.84) .. (6.56,1.97)   ;
\draw    (591.82,168.12) -- (581.52,157.82) ;
\draw [shift={(580.11,156.41)}, rotate = 45] [color={rgb, 255:red, 0; green, 0; blue, 0 }  ][line width=0.75]    (6.56,-1.97) .. controls (4.17,-0.84) and (1.99,-0.18) .. (0,0) .. controls (1.99,0.18) and (4.17,0.84) .. (6.56,1.97)   ;
\draw  [color={rgb, 255:red, 255; green, 174; blue, 10 }  ,draw opacity=1 ][dash pattern={on 4.5pt off 4.5pt}] (414.75,155.8) .. controls (414.75,147.63) and (421.38,141) .. (429.56,141) -- (486.3,141) .. controls (494.48,141) and (501.11,147.63) .. (501.11,155.8) -- (501.11,200.22) .. controls (501.11,208.39) and (494.48,215.02) .. (486.3,215.02) -- (429.56,215.02) .. controls (421.38,215.02) and (414.75,208.39) .. (414.75,200.22) -- cycle ;
\draw  [dash pattern={on 4.5pt off 4.5pt}]  (377.27,177.69) -- (411.05,177.69) ;
\draw [shift={(413.05,177.69)}, rotate = 180] [color={rgb, 255:red, 0; green, 0; blue, 0 }  ][line width=0.75]    (8.74,-2.63) .. controls (5.56,-1.12) and (2.65,-0.24) .. (0,0) .. controls (2.65,0.24) and (5.56,1.12) .. (8.74,2.63)   ;
\draw  [dash pattern={on 4.5pt off 4.5pt}]  (502.8,177.7) -- (536.57,177.7) ;
\draw [shift={(538.57,177.7)}, rotate = 180] [color={rgb, 255:red, 0; green, 0; blue, 0 }  ][line width=0.75]    (8.74,-2.63) .. controls (5.56,-1.12) and (2.65,-0.24) .. (0,0) .. controls (2.65,0.24) and (5.56,1.12) .. (8.74,2.63)   ;
\draw    (576.61,166.68) -- (576.61,159.77) ;
\draw [shift={(576.61,157.77)}, rotate = 90] [color={rgb, 255:red, 0; green, 0; blue, 0 }  ][line width=0.75]    (6.56,-1.97) .. controls (4.17,-0.84) and (1.99,-0.18) .. (0,0) .. controls (1.99,0.18) and (4.17,0.84) .. (6.56,1.97)   ;

\draw (293.82,197.72) node [anchor=north west][inner sep=0.75pt]  [font=\normalsize] [align=left] {\begin{minipage}[lt]{54.57pt}\setlength\topsep{0pt}
\begin{center}
{\fontfamily{pcr}\selectfont \textbf{{\scriptsize Code Problem}}}
\end{center}

\end{minipage}};
\draw (425.55,177.06) node [anchor=north west][inner sep=0.75pt]  [font=\normalsize] [align=left] {\begin{minipage}[lt]{44.88pt}\setlength\topsep{0pt}
\begin{center}
{\fontfamily{pcr}\selectfont \textbf{{\footnotesize Generalist}}}\\{\fontfamily{pcr}\selectfont \textbf{{\footnotesize Agent}}}
\end{center}

\end{minipage}};
\draw (544.56,197.72) node [anchor=north west][inner sep=0.75pt]  [font=\normalsize] [align=left] {\begin{minipage}[lt]{55.15pt}\setlength\topsep{0pt}
\begin{center}
{\fontfamily{pcr}\selectfont \textbf{{\scriptsize Problems Tree}}}
\end{center}

\end{minipage}};

\end{tikzpicture}

\caption{Phase 1: Problems tree initialization.}
\label{fig:phase1}
\end{figure}
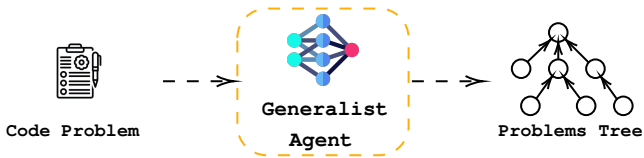

\par 
The framework begins with a Generalist Agent that performs recursive problem decomposition, as illustrated in Figure \ref{fig:phase1}. Given a complex coding task $X$, the agent:

\begin{enumerate}
    \item Decomposes the problem into constituent functions or code parts necessary for the task.
    \item Continues this decomposition recursively until reaching atomic units.
    \item Creates a tree structure where:
    \begin{itemize}
        \item The root node represents the overall problem or task.
        \item Internal nodes represent intermediate sub-problems.
        \item Leaf nodes represent practically indivisible coding tasks.
        \item Each node maintains specific functionality required by its parent solution.
    \end{itemize}
\end{enumerate}

\par 
This decomposition emphasizes extreme modularity by ensuring that every operation (including basic constructs like if-else or loop blocks) is encapsulated within documented, pure functions with highly descriptive names. The approach draws from functional programming principles, prioritizing ease of compositionality, modularity, expressiveness, and high readability.

\subsection{Bottom-up Code Generation}

\par 
Once the problem tree is established, code generation proceeds from leaves to root through the following steps:

\begin{enumerate}
    \item \textbf{Leaf Node Resolution}
    \begin{itemize}
        \item A Code Agent generates single, self-contained functions or other programming language constructs for each leaf node.
        \item Each function undergoes immediate testing and validation.
        \item Solutions employ chain-of-thought reasoning and incorporate feedback from tests.
    \end{itemize}

\begin{figure}[ht]
\centering

\tikzset{every picture/.style={line width=0.75pt}} 

\begin{tikzpicture}[x=0.75pt,y=0.75pt,yscale=-1,xscale=1]

\draw (382.38,117.19) node  {\includegraphics[width=25.06pt,height=25.06pt]{7747363.png}};
\draw   (190.6,98.73) .. controls (190.6,96.39) and (192.5,94.49) .. (194.83,94.49) .. controls (197.17,94.49) and (199.07,96.39) .. (199.07,98.73) .. controls (199.07,101.07) and (197.17,102.96) .. (194.83,102.96) .. controls (192.5,102.96) and (190.6,101.07) .. (190.6,98.73) -- cycle ;
\draw   (207.54,115.24) .. controls (207.54,112.9) and (209.43,111.01) .. (211.77,111.01) .. controls (214.11,111.01) and (216,112.9) .. (216,115.24) .. controls (216,117.58) and (214.11,119.47) .. (211.77,119.47) .. controls (209.43,119.47) and (207.54,117.58) .. (207.54,115.24) -- cycle ;
\draw   (190.6,115.24) .. controls (190.6,112.9) and (192.5,111.01) .. (194.83,111.01) .. controls (197.17,111.01) and (199.07,112.9) .. (199.07,115.24) .. controls (199.07,117.58) and (197.17,119.47) .. (194.83,119.47) .. controls (192.5,119.47) and (190.6,117.58) .. (190.6,115.24) -- cycle ;
\draw   (173.66,115.24) .. controls (173.66,112.9) and (175.56,111.01) .. (177.9,111.01) .. controls (180.24,111.01) and (182.13,112.9) .. (182.13,115.24) .. controls (182.13,117.58) and (180.24,119.47) .. (177.9,119.47) .. controls (175.56,119.47) and (173.66,117.58) .. (173.66,115.24) -- cycle ;
\draw   (220.24,132.18) .. controls (220.24,129.84) and (222.13,127.94) .. (224.47,127.94) .. controls (226.81,127.94) and (228.71,129.84) .. (228.71,132.18) .. controls (228.71,134.51) and (226.81,136.41) .. (224.47,136.41) .. controls (222.13,136.41) and (220.24,134.51) .. (220.24,132.18) -- cycle ;
\draw   (177.9,132.18) .. controls (177.9,129.84) and (179.79,127.94) .. (182.13,127.94) .. controls (184.47,127.94) and (186.37,129.84) .. (186.37,132.18) .. controls (186.37,134.51) and (184.47,136.41) .. (182.13,136.41) .. controls (179.79,136.41) and (177.9,134.51) .. (177.9,132.18) -- cycle ;
\draw   (203.3,132.18) .. controls (203.3,129.84) and (205.2,127.94) .. (207.54,127.94) .. controls (209.87,127.94) and (211.77,129.84) .. (211.77,132.18) .. controls (211.77,134.51) and (209.87,136.41) .. (207.54,136.41) .. controls (205.2,136.41) and (203.3,134.51) .. (203.3,132.18) -- cycle ;
\draw    (184.87,128.82) -- (190.91,120.29) ;
\draw [shift={(192.07,118.66)}, rotate = 125.31] [color={rgb, 255:red, 0; green, 0; blue, 0 }  ][line width=0.75]    (6.56,-1.97) .. controls (4.17,-0.84) and (1.99,-0.18) .. (0,0) .. controls (1.99,0.18) and (4.17,0.84) .. (6.56,1.97)   ;
\draw    (204.77,128.82) -- (198.35,120.26) ;
\draw [shift={(197.15,118.66)}, rotate = 53.13] [color={rgb, 255:red, 0; green, 0; blue, 0 }  ][line width=0.75]    (6.56,-1.97) .. controls (4.17,-0.84) and (1.99,-0.18) .. (0,0) .. controls (1.99,0.18) and (4.17,0.84) .. (6.56,1.97)   ;
\draw    (180.66,112.24) -- (189.99,103.32) ;
\draw [shift={(191.43,101.94)}, rotate = 136.28] [color={rgb, 255:red, 0; green, 0; blue, 0 }  ][line width=0.75]    (6.56,-1.97) .. controls (4.17,-0.84) and (1.99,-0.18) .. (0,0) .. controls (1.99,0.18) and (4.17,0.84) .. (6.56,1.97)   ;
\draw    (221.5,129.04) -- (215.25,120.29) ;
\draw [shift={(214.09,118.66)}, rotate = 54.46] [color={rgb, 255:red, 0; green, 0; blue, 0 }  ][line width=0.75]    (6.56,-1.97) .. controls (4.17,-0.84) and (1.99,-0.18) .. (0,0) .. controls (1.99,0.18) and (4.17,0.84) .. (6.56,1.97)   ;
\draw    (208.58,112.31) -- (199.41,103.14) ;
\draw [shift={(198,101.73)}, rotate = 45] [color={rgb, 255:red, 0; green, 0; blue, 0 }  ][line width=0.75]    (6.56,-1.97) .. controls (4.17,-0.84) and (1.99,-0.18) .. (0,0) .. controls (1.99,0.18) and (4.17,0.84) .. (6.56,1.97)   ;
\draw  [color={rgb, 255:red, 255; green, 174; blue, 10 }  ,draw opacity=1 ][dash pattern={on 4.5pt off 4.5pt}] (342.96,101.6) .. controls (342.96,94.21) and (348.95,88.22) .. (356.34,88.22) -- (407.62,88.22) .. controls (415.01,88.22) and (421,94.21) .. (421,101.6) -- (421,141.73) .. controls (421,149.12) and (415.01,155.11) .. (407.62,155.11) -- (356.34,155.11) .. controls (348.95,155.11) and (342.96,149.12) .. (342.96,141.73) -- cycle ;
\draw  [dash pattern={on 4.5pt off 4.5pt}]  (278.7,121.38) -- (336,121.63) ;
\draw [shift={(338,121.64)}, rotate = 180.25] [color={rgb, 255:red, 0; green, 0; blue, 0 }  ][line width=0.75]    (8.74,-2.63) .. controls (5.56,-1.12) and (2.65,-0.24) .. (0,0) .. controls (2.65,0.24) and (5.56,1.12) .. (8.74,2.63)   ;
\draw  [dash pattern={on 4.5pt off 4.5pt}]  (423.53,121.39) -- (453.85,121.39) ;
\draw [shift={(455.85,121.39)}, rotate = 180] [color={rgb, 255:red, 0; green, 0; blue, 0 }  ][line width=0.75]    (8.74,-2.63) .. controls (5.56,-1.12) and (2.65,-0.24) .. (0,0) .. controls (2.65,0.24) and (5.56,1.12) .. (8.74,2.63)   ;
\draw    (194.83,111.01) -- (194.83,104.96) ;
\draw [shift={(194.83,102.96)}, rotate = 90] [color={rgb, 255:red, 0; green, 0; blue, 0 }  ][line width=0.75]    (6.56,-1.97) .. controls (4.17,-0.84) and (1.99,-0.18) .. (0,0) .. controls (1.99,0.18) and (4.17,0.84) .. (6.56,1.97)   ;
\draw  [color={rgb, 255:red, 0; green, 0; blue, 0 }  ,draw opacity=1 ][dash pattern={on 4.5pt off 4.5pt}] (174.33,128.51) .. controls (174.33,127.12) and (175.45,126) .. (176.83,126) -- (230.3,126) .. controls (231.69,126) and (232.81,127.12) .. (232.81,128.51) -- (232.81,136.03) .. controls (232.81,137.41) and (231.69,138.53) .. (230.3,138.53) -- (176.83,138.53) .. controls (175.45,138.53) and (174.33,137.41) .. (174.33,136.03) -- cycle ;
\draw  [color={rgb, 255:red, 0; green, 0; blue, 0 }  ,draw opacity=1 ][dash pattern={on 4.5pt off 4.5pt}] (170.53,111.48) .. controls (170.53,110.1) and (171.65,108.97) .. (173.03,108.97) -- (182.76,108.97) .. controls (184.15,108.97) and (185.27,110.1) .. (185.27,111.48) -- (185.27,119) .. controls (185.27,120.38) and (184.15,121.51) .. (182.76,121.51) -- (173.03,121.51) .. controls (171.65,121.51) and (170.53,120.38) .. (170.53,119) -- cycle ;
\draw  [dash pattern={on 4.5pt off 4.5pt}]  (233.09,131.34) .. controls (247.96,122.56) and (244.12,121.89) .. (278.7,121.38) ;
\draw  [dash pattern={on 4.5pt off 4.5pt}]  (177.36,108.96) .. controls (177.28,79.28) and (179.81,84.11) .. (220.47,83.99) .. controls (261.13,83.88) and (244.23,121.98) .. (269.66,121.61) ;
\draw (477.79,121.22) node  {\includegraphics[width=25.06pt,height=25.06pt]{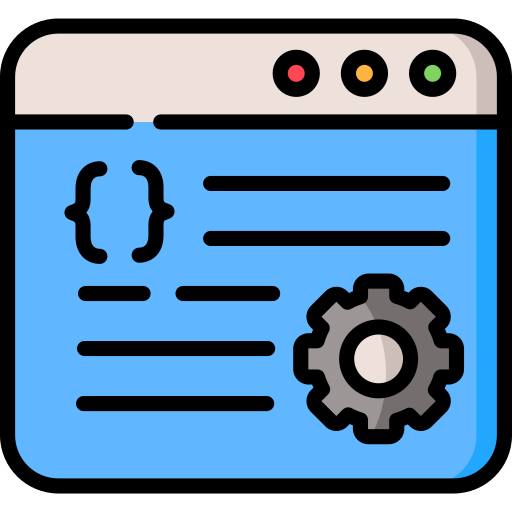}};
\draw (381.7,29.95) node  {\includegraphics[width=31.33pt,height=31.33pt]{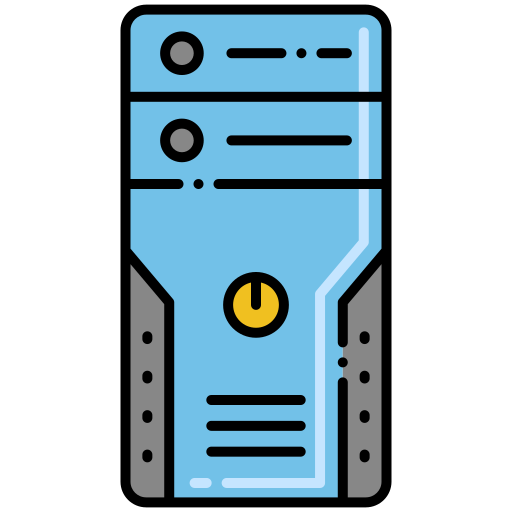}};
\draw  [color={rgb, 255:red, 255; green, 174; blue, 10 }  ,draw opacity=1 ][dash pattern={on 4.5pt off 4.5pt}] (342.98,16.43) .. controls (342.98,9.04) and (348.97,3.05) .. (356.36,3.05) -- (407.64,3.05) .. controls (415.03,3.05) and (421.02,9.04) .. (421.02,16.43) -- (421.02,56.56) .. controls (421.02,63.95) and (415.03,69.94) .. (407.64,69.94) -- (356.36,69.94) .. controls (348.97,69.94) and (342.98,63.95) .. (342.98,56.56) -- cycle ;
\draw  [dash pattern={on 4.5pt off 4.5pt}]  (477,100.75) .. controls (477.05,92.61) and (477,67.57) .. (476.96,56.26) .. controls (476.92,44.94) and (466.43,35.5) .. (452.17,35.61) .. controls (439.41,35.71) and (432.42,35.81) .. (426.8,35.65) ;
\draw [shift={(424.87,35.59)}, rotate = 2.27] [color={rgb, 255:red, 0; green, 0; blue, 0 }  ][line width=0.75]    (8.74,-2.63) .. controls (5.56,-1.12) and (2.65,-0.24) .. (0,0) .. controls (2.65,0.24) and (5.56,1.12) .. (8.74,2.63)   ;
\draw (287.55,35.6) node  {\includegraphics[width=15.66pt,height=15.66pt]{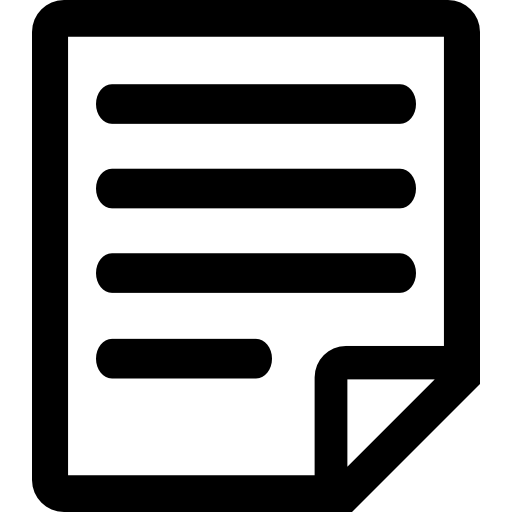}};
\draw  [dash pattern={on 4.5pt off 4.5pt}]  (339.89,35.59) -- (303.81,35.59) ;
\draw [shift={(301.81,35.59)}, rotate = 360] [color={rgb, 255:red, 0; green, 0; blue, 0 }  ][line width=0.75]    (8.74,-2.63) .. controls (5.56,-1.12) and (2.65,-0.24) .. (0,0) .. controls (2.65,0.24) and (5.56,1.12) .. (8.74,2.63)   ;
\draw  [dash pattern={on 4.5pt off 4.5pt}]  (287.15,63.16) .. controls (287.15,73.09) and (286.78,85.53) .. (287.2,96.81) .. controls (287.62,108.09) and (296.07,113.46) .. (307.99,113.28) .. controls (319.02,113.11) and (327.82,113.26) .. (336.02,113.28) ;
\draw [shift={(338,113.29)}, rotate = 180] [color={rgb, 255:red, 0; green, 0; blue, 0 }  ][line width=0.75]    (8.74,-2.63) .. controls (5.56,-1.12) and (2.65,-0.24) .. (0,0) .. controls (2.65,0.24) and (5.56,1.12) .. (8.74,2.63)   ;

\draw (349.83,138.87) node [anchor=north west][inner sep=0.75pt]  [font=\normalsize] [align=left] {\begin{minipage}[lt]{45.6pt}\setlength\topsep{0pt}
\begin{center}
{\scriptsize {\fontfamily{pcr}\selectfont \textbf{Code Agent}}}
\end{center}

\end{minipage}};
\draw (162.84,143.28) node [anchor=north west][inner sep=0.75pt]  [font=\normalsize] [align=left] {\begin{minipage}[lt]{55.13pt}\setlength\topsep{0pt}
\begin{center}
{\scriptsize {\fontfamily{pcr}\selectfont \textbf{Problems Tree}}}
\end{center}

\end{minipage}};
\draw (260.65,126.86) node [anchor=north west][inner sep=0.75pt]   [align=left] {{\scriptsize {\fontfamily{pcr}\selectfont \textbf{Leaf Problems}}}};
\draw (437.73,140.8) node [anchor=north west][inner sep=0.75pt]   [align=left] {{\scriptsize {\fontfamily{pcr}\selectfont \textbf{Code Solution}}}};
\draw (346.63,54.43) node [anchor=north west][inner sep=0.75pt]   [align=left] {{\scriptsize {\fontfamily{pcr}\selectfont \textbf{Tester Agent}}}};
\draw (264.14,49.72) node [anchor=north west][inner sep=0.75pt]   [align=left] {{\scriptsize {\fontfamily{pcr}\selectfont \textbf{Feedback}}}};

\end{tikzpicture}

\caption{Phase 2: Solving leaf problems.}
\end{figure}

\item \textbf{Upward Composition}
\begin{figure}[ht]
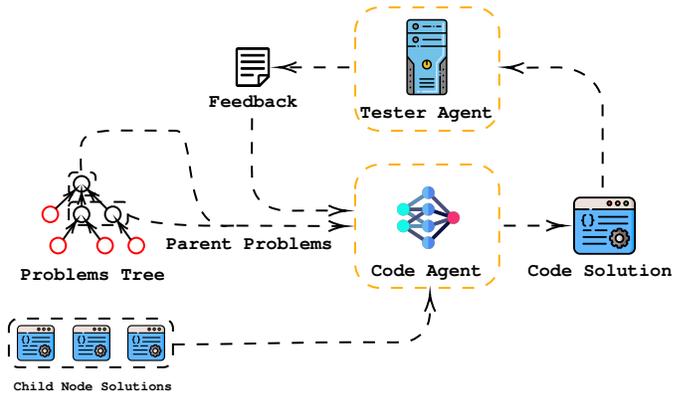

\centering

\tikzset{every picture/.style={line width=0.75pt}} 

\begin{tikzpicture}[x=0.75pt,y=0.75pt,yscale=-1,xscale=1]

\draw (391.51,109.35) node  {\includegraphics[width=23.34pt,height=23.34pt]{7747363.png}};
\draw   (212.91,92.16) .. controls (212.91,89.98) and (214.67,88.21) .. (216.85,88.21) .. controls (219.03,88.21) and (220.79,89.98) .. (220.79,92.16) .. controls (220.79,94.33) and (219.03,96.1) .. (216.85,96.1) .. controls (214.67,96.1) and (212.91,94.33) .. (212.91,92.16) -- cycle ;
\draw   (228.68,107.54) .. controls (228.68,105.36) and (230.44,103.59) .. (232.62,103.59) .. controls (234.8,103.59) and (236.56,105.36) .. (236.56,107.54) .. controls (236.56,109.71) and (234.8,111.48) .. (232.62,111.48) .. controls (230.44,111.48) and (228.68,109.71) .. (228.68,107.54) -- cycle ;
\draw   (212.91,107.54) .. controls (212.91,105.36) and (214.67,103.59) .. (216.85,103.59) .. controls (219.03,103.59) and (220.79,105.36) .. (220.79,107.54) .. controls (220.79,109.71) and (219.03,111.48) .. (216.85,111.48) .. controls (214.67,111.48) and (212.91,109.71) .. (212.91,107.54) -- cycle ;
\draw  [color={rgb, 255:red, 255; green, 0; blue, 0 }  ,draw opacity=1 ] (197.13,107.54) .. controls (197.13,105.36) and (198.9,103.59) .. (201.08,103.59) .. controls (203.25,103.59) and (205.02,105.36) .. (205.02,107.54) .. controls (205.02,109.71) and (203.25,111.48) .. (201.08,111.48) .. controls (198.9,111.48) and (197.13,109.71) .. (197.13,107.54) -- cycle ;
\draw  [color={rgb, 255:red, 255; green, 0; blue, 0 }  ,draw opacity=1 ] (240.51,123.31) .. controls (240.51,121.13) and (242.27,119.36) .. (244.45,119.36) .. controls (246.63,119.36) and (248.39,121.13) .. (248.39,123.31) .. controls (248.39,125.49) and (246.63,127.25) .. (244.45,127.25) .. controls (242.27,127.25) and (240.51,125.49) .. (240.51,123.31) -- cycle ;
\draw  [color={rgb, 255:red, 255; green, 0; blue, 0 }  ,draw opacity=1 ] (201.08,123.31) .. controls (201.08,121.13) and (202.84,119.36) .. (205.02,119.36) .. controls (207.2,119.36) and (208.96,121.13) .. (208.96,123.31) .. controls (208.96,125.49) and (207.2,127.25) .. (205.02,127.25) .. controls (202.84,127.25) and (201.08,125.49) .. (201.08,123.31) -- cycle ;
\draw  [color={rgb, 255:red, 255; green, 0; blue, 0 }  ,draw opacity=1 ] (224.74,123.31) .. controls (224.74,121.13) and (226.5,119.36) .. (228.68,119.36) .. controls (230.86,119.36) and (232.62,121.13) .. (232.62,123.31) .. controls (232.62,125.49) and (230.86,127.25) .. (228.68,127.25) .. controls (226.5,127.25) and (224.74,125.49) .. (224.74,123.31) -- cycle ;
\draw    (207.57,120.19) -- (213.12,112.35) ;
\draw [shift={(214.27,110.72)}, rotate = 125.31] [color={rgb, 255:red, 0; green, 0; blue, 0 }  ][line width=0.75]    (6.56,-1.97) .. controls (4.17,-0.84) and (1.99,-0.18) .. (0,0) .. controls (1.99,0.18) and (4.17,0.84) .. (6.56,1.97)   ;
\draw    (226.1,120.19) -- (220.21,112.32) ;
\draw [shift={(219.01,110.72)}, rotate = 53.13] [color={rgb, 255:red, 0; green, 0; blue, 0 }  ][line width=0.75]    (6.56,-1.97) .. controls (4.17,-0.84) and (1.99,-0.18) .. (0,0) .. controls (1.99,0.18) and (4.17,0.84) .. (6.56,1.97)   ;
\draw    (203.65,104.74) -- (212.24,96.53) ;
\draw [shift={(213.68,95.15)}, rotate = 136.28] [color={rgb, 255:red, 0; green, 0; blue, 0 }  ][line width=0.75]    (6.56,-1.97) .. controls (4.17,-0.84) and (1.99,-0.18) .. (0,0) .. controls (1.99,0.18) and (4.17,0.84) .. (6.56,1.97)   ;
\draw    (241.68,120.38) -- (235.94,112.35) ;
\draw [shift={(234.78,110.72)}, rotate = 54.46] [color={rgb, 255:red, 0; green, 0; blue, 0 }  ][line width=0.75]    (6.56,-1.97) .. controls (4.17,-0.84) and (1.99,-0.18) .. (0,0) .. controls (1.99,0.18) and (4.17,0.84) .. (6.56,1.97)   ;
\draw    (229.65,104.81) -- (221.21,96.36) ;
\draw [shift={(219.8,94.95)}, rotate = 45] [color={rgb, 255:red, 0; green, 0; blue, 0 }  ][line width=0.75]    (6.56,-1.97) .. controls (4.17,-0.84) and (1.99,-0.18) .. (0,0) .. controls (1.99,0.18) and (4.17,0.84) .. (6.56,1.97)   ;
\draw  [color={rgb, 255:red, 255; green, 174; blue, 10 }  ,draw opacity=1 ][dash pattern={on 4.5pt off 4.5pt}] (354.8,94.83) .. controls (354.8,87.95) and (360.38,82.37) .. (367.26,82.37) -- (415.02,82.37) .. controls (421.9,82.37) and (427.48,87.95) .. (427.48,94.83) -- (427.48,132.21) .. controls (427.48,139.09) and (421.9,144.67) .. (415.02,144.67) -- (367.26,144.67) .. controls (360.38,144.67) and (354.8,139.09) .. (354.8,132.21) -- cycle ;
\draw  [dash pattern={on 4.5pt off 4.5pt}]  (294.96,113.25) -- (348.18,113.49) ;
\draw [shift={(350.18,113.5)}, rotate = 180.25] [color={rgb, 255:red, 0; green, 0; blue, 0 }  ][line width=0.75]    (8.74,-2.63) .. controls (5.56,-1.12) and (2.65,-0.24) .. (0,0) .. controls (2.65,0.24) and (5.56,1.12) .. (8.74,2.63)   ;
\draw  [dash pattern={on 4.5pt off 4.5pt}]  (429.83,113.26) -- (457.94,113.26) ;
\draw [shift={(459.94,113.26)}, rotate = 180] [color={rgb, 255:red, 0; green, 0; blue, 0 }  ][line width=0.75]    (8.74,-2.63) .. controls (5.56,-1.12) and (2.65,-0.24) .. (0,0) .. controls (2.65,0.24) and (5.56,1.12) .. (8.74,2.63)   ;
\draw    (216.85,103.59) -- (216.85,98.1) ;
\draw [shift={(216.85,96.1)}, rotate = 90] [color={rgb, 255:red, 0; green, 0; blue, 0 }  ][line width=0.75]    (6.56,-1.97) .. controls (4.17,-0.84) and (1.99,-0.18) .. (0,0) .. controls (1.99,0.18) and (4.17,0.84) .. (6.56,1.97)   ;
\draw  [color={rgb, 255:red, 0; green, 0; blue, 0 }  ,draw opacity=1 ][dash pattern={on 4.5pt off 4.5pt}] (210.39,103.59) .. controls (210.39,102.3) and (211.44,101.26) .. (212.73,101.26) -- (237.86,101.26) .. controls (239.15,101.26) and (240.19,102.3) .. (240.19,103.59) -- (240.19,110.59) .. controls (240.19,111.88) and (239.15,112.93) .. (237.86,112.93) -- (212.73,112.93) .. controls (211.44,112.93) and (210.39,111.88) .. (210.39,110.59) -- cycle ;
\draw  [dash pattern={on 4.5pt off 4.5pt}]  (240.37,107.82) .. controls (246.8,111.54) and (262.75,113.73) .. (294.96,113.25) ;
\draw  [dash pattern={on 4.5pt off 4.5pt}]  (216.58,85.4) .. controls (216.33,79.64) and (216.46,81.83) .. (216.46,74.54) .. controls (216.46,67.24) and (222.48,65.21) .. (235.65,65.21) .. controls (248.82,65.21) and (253.11,65.3) .. (264.24,65.33) .. controls (275.38,65.36) and (266.26,113.82) .. (289.95,113.47) ;
\draw (480.37,113.11) node  {\includegraphics[width=23.34pt,height=23.34pt]{3662830.png}};
\draw (390.89,28.1) node  {\includegraphics[width=29.18pt,height=29.18pt]{3103969.png}};
\draw  [color={rgb, 255:red, 255; green, 174; blue, 10 }  ,draw opacity=1 ][dash pattern={on 4.5pt off 4.5pt}] (354.82,15.51) .. controls (354.82,8.63) and (360.4,3.05) .. (367.28,3.05) -- (415.04,3.05) .. controls (421.92,3.05) and (427.5,8.63) .. (427.5,15.51) -- (427.5,52.89) .. controls (427.5,59.77) and (421.92,65.35) .. (415.04,65.35) -- (367.28,65.35) .. controls (360.4,65.35) and (354.82,59.77) .. (354.82,52.89) -- cycle ;
\draw  [dash pattern={on 4.5pt off 4.5pt}]  (479.64,94.04) .. controls (479.68,86.46) and (479.64,63.14) .. (479.6,52.6) .. controls (479.56,42.07) and (469.79,33.27) .. (456.51,33.37) .. controls (444.76,33.47) and (438.26,33.55) .. (433.06,33.42) ;
\draw [shift={(431.09,33.36)}, rotate = 2.27] [color={rgb, 255:red, 0; green, 0; blue, 0 }  ][line width=0.75]    (8.74,-2.63) .. controls (5.56,-1.12) and (2.65,-0.24) .. (0,0) .. controls (2.65,0.24) and (5.56,1.12) .. (8.74,2.63)   ;
\draw (303.2,33.36) node  {\includegraphics[width=14.59pt,height=14.59pt]{45503.png}};
\draw  [dash pattern={on 4.5pt off 4.5pt}]  (351.94,33.36) -- (318.48,33.36) ;
\draw [shift={(316.48,33.36)}, rotate = 360] [color={rgb, 255:red, 0; green, 0; blue, 0 }  ][line width=0.75]    (8.74,-2.63) .. controls (5.56,-1.12) and (2.65,-0.24) .. (0,0) .. controls (2.65,0.24) and (5.56,1.12) .. (8.74,2.63)   ;
\draw  [dash pattern={on 4.5pt off 4.5pt}]  (302.82,59.03) .. controls (302.82,68.28) and (302.48,79.87) .. (302.87,90.37) .. controls (303.26,100.88) and (311.13,105.87) .. (322.23,105.71) .. controls (332.45,105.55) and (340.61,105.69) .. (348.21,105.71) ;
\draw [shift={(350.18,105.71)}, rotate = 180] [color={rgb, 255:red, 0; green, 0; blue, 0 }  ][line width=0.75]    (8.74,-2.63) .. controls (5.56,-1.12) and (2.65,-0.24) .. (0,0) .. controls (2.65,0.24) and (5.56,1.12) .. (8.74,2.63)   ;
\draw  [color={rgb, 255:red, 0; green, 0; blue, 0 }  ,draw opacity=1 ][dash pattern={on 4.5pt off 4.5pt}] (210.48,88.69) .. controls (210.48,87.4) and (211.53,86.36) .. (212.82,86.36) -- (221.56,86.36) .. controls (222.85,86.36) and (223.9,87.4) .. (223.9,88.69) -- (223.9,95.69) .. controls (223.9,96.98) and (222.85,98.03) .. (221.56,98.03) -- (212.82,98.03) .. controls (211.53,98.03) and (210.48,96.98) .. (210.48,95.69) -- cycle ;
\draw (221.66,172.5) node  {\includegraphics[width=13.97pt,height=13.97pt]{3662830.png}};
\draw (249.6,172.5) node  {\includegraphics[width=13.97pt,height=13.97pt]{3662830.png}};
\draw (193.72,172.5) node  {\includegraphics[width=13.97pt,height=13.97pt]{3662830.png}};
\draw  [dash pattern={on 4.5pt off 4.5pt}] (180.17,165.06) .. controls (180.17,162.35) and (182.37,160.14) .. (185.09,160.14) -- (258.03,160.14) .. controls (260.74,160.14) and (262.94,162.35) .. (262.94,165.06) -- (262.94,179.81) .. controls (262.94,182.53) and (260.74,184.73) .. (258.03,184.73) -- (185.09,184.73) .. controls (182.37,184.73) and (180.17,182.53) .. (180.17,179.81) -- cycle ;
\draw  [dash pattern={on 4.5pt off 4.5pt}]  (262.87,173.22) .. controls (286.88,173.2) and (334.2,172.02) .. (357.19,172.06) .. controls (380.19,172.11) and (381.7,173.03) .. (387.63,168.42) .. controls (393.09,164.18) and (392.71,157.06) .. (392.6,149.76) ;
\draw [shift={(392.58,147.85)}, rotate = 89.88] [color={rgb, 255:red, 0; green, 0; blue, 0 }  ][line width=0.75]    (8.74,-2.63) .. controls (5.56,-1.12) and (2.65,-0.24) .. (0,0) .. controls (2.65,0.24) and (5.56,1.12) .. (8.74,2.63)   ;

\draw (359,130.99) node [anchor=north west][inner sep=0.75pt]  [font=\normalsize] [align=left] {\begin{minipage}[lt]{45.6pt}\setlength\topsep{0pt}
\begin{center}
{\scriptsize {\fontfamily{pcr}\selectfont \textbf{Code Agent}}}
\end{center}

\end{minipage}};
\draw (184.37,133.1) node [anchor=north west][inner sep=0.75pt]  [font=\normalsize] [align=left] {\begin{minipage}[lt]{55.13pt}\setlength\topsep{0pt}
\begin{center}
{\scriptsize {\fontfamily{pcr}\selectfont \textbf{Problems Tree}}}
\end{center}

\end{minipage}};
\draw (258.05,117.81) node [anchor=north west][inner sep=0.75pt]   [align=left] {{\scriptsize {\fontfamily{pcr}\selectfont \textbf{Parent Problems}}}};
\draw (440.46,130.79) node [anchor=north west][inner sep=0.75pt]   [align=left] {{\scriptsize {\fontfamily{pcr}\selectfont \textbf{Code Solution}}}};
\draw (355.82,52.22) node [anchor=north west][inner sep=0.75pt]   [align=left] {{\scriptsize {\fontfamily{pcr}\selectfont \textbf{Tester Agent}}}};
\draw (279.68,45.96) node [anchor=north west][inner sep=0.75pt]   [align=left] {{\scriptsize {\fontfamily{pcr}\selectfont \textbf{Feedback}}}};
\draw (181.14,190.42) node [anchor=north west][inner sep=0.75pt]   [align=left] {{\fontfamily{pcr}\selectfont \textbf{{\tiny Child Node Solutions}}}};

\end{tikzpicture}

\caption{Phase 3: Composing child nodes to solve parent node.}
\label{fig:framework}

\end{figure}

    \begin{itemize}
        \item For each non-leaf node, the Code Agent receives:
        \begin{itemize}
            \item A list of documentation statements for available functions or code blocks from child nodes.
            \item Function documentation and descriptions.
            \item No access to implementation details or problem descriptions of descendant nodes.
        \end{itemize}
        \item The agent composes solutions using only the provided function interfaces or code blocks.
        \item This process continues until reaching the root node.
    \end{itemize}

\end{enumerate}

\subsection{Multi-Agent Validation System}

\par 
Using the Chain-of-Thought prompting technique \cite{cot}, each problem is passed to the Generalist Agent to give a candidate solution, break it down, and determine the tools or functions needed to implement the task. Each generated solution undergoes rigorous validation through a comprehensive multi-agent evaluation process. The primary validation mechanism employs a dedicated Critic Agent that conducts thorough analyses of each solution. This agent performs detailed reviews focusing on multiple dimensions of code quality, including implementation efficiency and functional correctness. The feedback generated through this critical analysis is systematically incorporated into subsequent generation attempts, enabling continuous refinement of the solutions.

Furthermore, the validation process incorporates an automated testing phase executed by a specialized testing agent. This automated evaluation provides quantitative performance metrics and identifies potential issues in the generated code. The testing agent generates detailed diagnostic information and debugging suggestions, which are then integrated back into the generation process. This dual-agent validation approach ensures comprehensive quality assessment while maintaining a continuous feedback loop for solution optimization.

Although Chain of Thought (CoT) reasoning is recognized as an emergent capability primarily manifesting in larger language models—typically those exceeding 60-70 billion parameters—we found it necessary to implement CoT prompting with the smaller Llama 3.1 8B model to facilitate task planning and problem tree generation. Therefore, the observed performance boosts through is due to three key mechanisms: self-critique, problem decomposition, and upward composition. However, this approach introduces a critical vulnerability: if the initial proposed solution to the root problem is incorrect, this error propagates throughout the entire solution structure. Despite this limitation, our findings suggest that implementing this methodology on larger language models would likely yield superior performance due to their inherent CoT capabilities.

\subsection{Key Advantages}

Our framework addresses three key limitations of direct LLM-based code generation. First, it improves context window management by isolating sub-problems and providing specific context at each generation phase, thereby circumventing standard token limitations while maintaining solution coherence.

The framework also enhances reasoning by constraining the scope of individual generation tasks. This division enables the LLM to process well-defined sub-problems rather than attempting to generate complete solutions in a single step, resulting in more reliable code generation.

Finally, the framework reduces errors through a multi-layered approach. It combines validation systems for early error detection, chain-of-thought reasoning, and iterative feedback loops. The bottom-up compositional structure ensures that complex solutions are built from verified components, improving overall reliability.

\subsection{Theoretical Framework}

Our approach reconceptualizes code generation as a dual-nature problem comprising two fundamental problems. The first problem frames code generation for atomic components (leaf nodes) as an information retrieval and fuzzy search problem. This formulation capitalizes on the extensive code corpus available in contemporary training datasets, enabling Large Language Models to leverage their demonstrated proficiency in pattern recognition and approximate matching. The effectiveness of this approach stems from the models' exposure to diverse programming patterns during training, facilitating the retrieval and adaptation of relevant code segments.

The second paradigm addresses the compositional aspect of code generation, focusing on the systematic integration of verified components. This composition-centric perspective emphasizes the importance of interface-based integration rather than implementation-specific details. By abstracting the composition process to the interface level, we establish a more robust and generalizable framework for component integration, mitigating the complexities typically associated with implementation-level composition. This theoretical decomposition allows us to leverage LLMs' strength in pattern matching for atomic components while providing a structured approach to addressing their known limitations in compositional reasoning.

\par 
This structured approach enables reliable generation of complex software systems while maintaining high code quality and correctness.

\section{Results}

\par 
We evaluated our guided code generation framework against traditional one-shot generation approaches using OpenAI's HumanEval benchmark \cite{humaneval}. The experiments were conducted using Meta's Llama 3.1 8B model with int4 precision quantization\cite{nousresearchllama3.1} due to computational resource constraints. Our evaluation focused on two primary metrics: Pass@1 scores and qualitative analysis of the generated solutions.

\subsection{Quantitative Analysis}
\begin{table}[h]
\centering
\begin{tabular}{lcc}
\hline
\textbf{Approach} & \textbf{Pass@1 (\%)} & \textbf{Relative Improvement} \\
\hline
Direct, One-shot Generation &45.4 & - \\
Our Framework & 56.2 & +23.79\% \\
\hline
\\
\end{tabular}
\caption{Performance comparison on HumanEval benchmark.}
\label{tab:results}
\end{table}
\par 
Our framework demonstrated substantial improvements over direct code generation approaches. As shown in Table\ref{tab:results}, the guided code generation framework achieved a Pass@1 score of 56.2\% on the HumanEval benchmark, representing a 23.79 percentage point improvement over the baseline one-shot generation approach, which scored 45.4\% Pass@1. This improvement is particularly noteworthy given our use of a relatively small (8B parameters) and quantized model, suggesting that the structured approach can effectively compensate for model size and precision limitations.

\subsection{Qualitative Analysis}

\par
Beyond the HumanEval benchmark, we tested our framework's capability on long, complex coding tasks like building a mathematical function evaluator. The framework successfully generated it with a complete lexer, a complete parser, a complete evaluation algorithm, and with comprehensive error handling throughout every component of the evaluator. On the other hand, frontier models like OpenAI's GPT-4o and Google's Gemini 1.5 Pro consistently either refused to do the task or generated much simpler evaluators.

\section{Conclusion}

This paper introduces a novel framework for guided code generation addressing key limitations in current LLM-based approaches. Our methodology, combining hierarchical decomposition, bottom-up generation, and multi-agent validation, demonstrates significant improvements in code generation capabilities, even with modest model architectures.

Empirical results show a 23.79\% improvement in Pass@1 scores on the HumanEval benchmark using an 8B parameter model. Our framework successfully generated complex software systems that larger frontier models typically refuse, highlighting the potential of decomposition-based approaches.

We propose a theoretical framework reconceptualizing code generation as a dual problem of information retrieval and compositional reasoning. While aligning with our empirical results, this framework remains hypothetical and requires further study.

Future work should focus on scaling to larger models, exploring diverse programming paradigms, optimizing decomposition and validation processes, and rigorously testing the theoretical model.

Our findings underscore the potential of structured, multi-agent approaches in overcoming current LLM limitations, paving the way for more sophisticated code generation systems that combine traditional software engineering principles with modern AI capabilities.

\end{document}